\pdfoutput=1

\documentclass[11pt]{article}
\usepackage{amssymb}
\usepackage{amsmath}

\usepackage[preprint]{acl}

\usepackage{times}
\usepackage{latexsym}

\usepackage[T1]{fontenc}

\usepackage[utf8]{inputenc}

\usepackage{microtype}

\usepackage{inconsolata}

\usepackage{graphicx}
\usepackage{booktabs}
\usepackage{arydshln} 
\title{Enhancing Small LLM Alignment through Margin-Based Objective Modifications under Resource Constraints}

\author{Daren Yao \\
  CEE \\
  \texttt{darenyao@andrew.cmu.edu} \\\And
  Jinsong Yuan \\
  CEE\\
  \texttt{jinsongy@andrew.cmu.edu} \\}

\author{
\textbf{Daren Yao} \and \textbf{Jinsong Yuan} \and \textbf{Ruike Chen} \\
Carnegie Mellon University \\
\texttt{\{darenyao, jinsongy, ruikec\}@andrew.cmu.edu}
}

\begin{document}
\maketitle
\begin{abstract}
Small large language models (LLMs) often face difficulties in aligning output to human preferences, particularly when operating under severe performance gaps.
In this work, we propose two lightweight DPO-based variants---Adaptive Margin-Sigmoid Loss and APO-hinge-zero---to better address underperformance scenarios by introducing margin-based objectives and selective update mechanisms.

Our APO-hinge-zero method, which combines hinge-induced hard-example mining with the chosen-focused optimization of APO-zero, achieves strong results.
In AlpacaEval, APO-hinge-zero improves the win rate by +2.0 points and the length-controlled win rate by +1.4 points compared to the APO-zero baseline.
In MT-Bench, our methods maintain competitive performance in diverse categories, particularly excelling in STEM and Humanities tasks.

These results demonstrate that simple modifications to preference-based objectives can significantly enhance small LLM alignment under resource constraints, offering a practical path toward more efficient deployment.
\end{abstract}

\section{Introduction}
Large language models (LLMs) have led to significant gains in natural language processing tasks, producing coherent and contextually apt text. Yet even state-of-the-art models often yield outputs that deviate from human preferences, motivating extensive research into alignment techniques ranging from reinforcement learning with human feedback (RLHF) to Direct Preference Optimization (DPO)~\citep{wang2024comprehensivesurveyllmalignment}.

Although alignment research has predominantly focused on large-scale models with tens of billions of parameters, many real-world applications, particularly those constrained by cost, computation, or privacy, require the deployment of much smaller models (under 1B parameters). Recent studies~\citep{chen2024rolesmallmodelsllm} highlight the strategic importance of small LLMs for low-latency inference, edge computing, and private environments. However, aligning small models remains especially challenging due to their limited capacity to faithfully represent complex user preferences.

In this project, we address the question: \emph{Can lightweight modifications to preference-based alignment objectives substantially improve small LLMs under resource constraints?} Motivated by the observation that standard DPO formulations may not fully accommodate the performance gaps typical of small LLMs, we explore novel objective variants that integrate margin-based mechanisms. Specifically, we introduce \textbf{Adaptive Margin-Sigmoid Loss} and \textbf{APO-hinge-zero} (with a Softplus variant), with the aim of improving alignment efficiency by:
\begin{itemize}
    \item combining stable log-sigmoid optimisation with explicit margin separation,
    \item implementing hard-example mining to sparsify updates and focus learning on difficult samples,
    \item improving robustness against noisy or ambiguous preference data through mild label smoothing.
\end{itemize}

Our methods are computationally inexpensive, easy to integrate into existing preference-optimization pipelines, and particularly suited for small-scale models. Preliminary results on AlpacaEval and MT-Bench demonstrate that margin-based objectives can meaningfully boost both win rates and length-controlled win rates without incurring additional complexity.

In essence, this work highlights the critical role of objective design in small model alignment and provides practical insights into scalable, cost-effective strategies for deploying aligned LLMs beyond massive cloud infrastructures.

\section{Related Works}

\subsection{DPO}

Direct Preference Optimization (DPO) is a lightweight alignment algorithm that bypasses the complexity of reinforcement learning by reparameterizing the RLHF objective into a simple classification loss. Instead of first fitting a separate reward model and then running PPO (or another RL algorithm), DPO shows that under a Bradley–Terry preference model, the optimal policy can be expressed in closed form as a function of the model’s own logits relative to a reference policy. By applying a change of variables, a loss is derived that directly increases the logarithmic probability gap between the preferred and preferred responses, weighted to prevent model collapse \cite{rafailov2024directpreferenceoptimizationlanguage}.

After this derivation, the DPO objective becomes:

\begin{equation}
    \mathcal{L}_{\text{DPO}}(\pi_\theta;\,\pi_{\mathrm{ref}})=-\mathbb{E}_{(x,y_w,y_l)\sim D} \\ 
\end{equation}
\[
\Bigl[\log\sigma\bigl(
\beta \ln\!\tfrac{\pi_\theta(y_w\mid x)}{\pi_{\mathrm{ref}}(y_w\mid x)}- \beta \ln\!\tfrac{\pi_\theta(y_l\mid x)}{\pi_{\mathrm{ref}}(y_l\mid x)}\bigr)\Bigr] \hspace{0.5em}
\]

where:
\begin{itemize}
  \item $\sigma(z)=1/(1+e^{-z})$ is the logistic function,
  \item $\pi_{\mathrm{ref}}$ is typically the supervised fine-tuned model,
  \item $\beta>0$ controls how strongly the model resists drifting from $\pi_{\mathrm{ref}}$.
\end{itemize}

In practice, preference data may contain noisy or ambiguous comparisons, making hard preference optimization suboptimal. In TRL \cite{vonwerra2022trl}, the DPO loss is actually the conservative DPO loss. Conservative DPO introduces a label smoothing factor $\epsilon \in [0, 0.5)$ into the DPO loss, balancing the confidence between the preferred and less preferred responses:
\begin{equation}
\mathcal{L}_{\mathrm{cDPO}}(\pi_\theta; \pi_{\mathrm{ref}}) = -\mathbb{E}_{(x, y_w, y_l) \sim D}
\end{equation}
\[\left[(1-\epsilon) \log \sigma(\beta \cdot \Delta r) + \epsilon \log \sigma(-\beta \cdot \Delta r)\right],\]
where $\Delta r = \ln\frac{\pi_\theta(y_w \mid x)}{\pi_{\mathrm{ref}}(y_w \mid x)} - \ln\frac{\pi_\theta(y_l \mid x)}{\pi_{\mathrm{ref}}(y_l \mid x)}$.
When $\epsilon=0$, the conservative DPO reduces to the original DPO objective (which is the default setting of TRL).


\subsection{DPO–Hinge}

While DPO’s logistic loss softly enforces a preference gap, DPO-Hinge replaces cross-entropy with a hinge-margin loss to guarantee a minimum margin $\alpha$ between preferred and dispreferred logits.  This can accelerate learning on borderline examples at the cost of an extra hyperparameter:
\begin{equation}
    \mathcal{L}_{\text{DPO-Hinge}}(\pi_\theta;\,\pi_{\mathrm{ref}})
 = \mathbb{E}_{(x,y_w,y_l)\sim D}
\end{equation}
\[
\Bigl[\max\Bigl(0,\;\alpha
- \Bigl(\beta \ln\tfrac{\pi_\theta(y_w\mid x)}{\pi_{\mathrm{ref}}(y_w\mid x)}
- \beta \ln\tfrac{\pi_\theta(y_l\mid x)}{\pi_{\mathrm{ref}}(y_l\mid x)}\Bigr)\Bigr)\Bigr],
\]
where $\alpha>0$ is the desired minimum log–ratio gap.

\subsection{AOT}

Alignment via Optimal Transport (AOT) is a distributional alignment method that enforces first-order stochastic dominance (FSD) of chosen (positive) samples over rejected (negative) samples at the distributional level. Specifically, given empirical samples of rewards from the chosen responses \(\{u_i\}_{i=1}^n\) and rejected responses \(\{v_i\}_{i=1}^n\), AOT constructs empirical measures:
\[
\hat{\mu}_U = \frac{1}{n}\sum_{i=1}^n\delta_{u_i}, \quad \hat{\mu}_V = \frac{1}{n}\sum_{i=1}^n\delta_{v_i}.
\]

The violation of stochastic dominance can be expressed as a one-dimensional optimal transport (OT) problem with convex surrogate loss \(h(\cdot)\):
\begin{equation}\label{eq:aot_ot_cost}
\mathrm{OT}_h(\hat{\mu}_U,\hat{\mu}_V) = \frac{1}{n}\sum_{i=1}^n h\bigl(u_{(i)} - v_{(i)}\bigr),
\end{equation}
where \((u_{(i)}, v_{(i)})\) represent sorted samples from the chosen and rejected distributions respectively. A typical choice for the surrogate loss \(h\) includes logistic or squared-hinge functions.  

The AOT fine-tuning objective is then defined as minimizing this OT cost over the policy parameter \(\theta\):
\begin{equation}\label{eq:aot_objective}
\min_\theta\;\mathrm{OT}_h\left((r\circ\pi_\theta)\sharp\mu_+,\,(r\circ\pi_{\text{ref}})\sharp\mu_-\right),
\end{equation}
which can be efficiently optimized through sorting-based computation, providing a stable optimization landscape~\cite{melnyk2024distributionalpreferencealignmentllms}.

\subsection{APO}

Anchored Preference Optimization (APO) is a family of contrastive alignment objectives designed to give explicit control over how the model’s likelihood of preferred and non-preferred responses is changed during training.  While DPO simply maximizes the gap between preferred and non-preferred log-probabilities, APO “anchors” each side of the comparison to either increase or decrease in a predictable way, depending on the relative quality of the examples and the target model\cite{doosterlinck2024anchoredpreferenceoptimizationcontrastive}.

Concretely, the following equation is the implicit reward (log-ratio) for response \(y\) under prompt \(x\).

\begin{equation}
    r_\theta(x,y) \;=\;\beta\;\ln\frac{\pi_\theta(y\mid x)}{\pi_{\mathrm{ref}}(y\mid x)}
\end{equation}

APO defines two core variants:

\paragraph{APO-zero} explicitly {\em increases} the likelihood of the preferred response and {\em decreases} that of the non-preferred response:

\begin{equation}
\mathcal{L}_{\text{APO-zero}}(x,y_w,y_l;\theta)= \hspace{4em}
\end{equation}
\[        
    \hspace{4em}-\sigma\!\bigl(r_\theta(x,y_w)\bigr)
    +\sigma\!\bigl(r_\theta(x,y_l)\bigr),
\]

\paragraph{APO-down} method {\em decreases} both likelihoods, but decreases the non-preferred more strongly:
\begin{equation}
    \mathcal{L}_{\text{APO-down}}(x,y_w,y_l;\theta)= \hspace{4em}
\end{equation}
\[
\hspace{4em}\sigma\bigl(r_\theta(x,y_w)\bigr)
-\sigma\bigl(r_\theta(x,y_w)-r_\theta(x,y_l)\bigr)
\]

Here \(\sigma(z)=1/(1+e^{-z})\) and \(\beta>0\) controls how far the model may drift from the reference policy 

\subsection{Other Alignment Methods}

Beyond DPO, several extensions such as IPO \citep{azar2023generaltheoreticalparadigmunderstand} and EXO \citep{ji2024efficientexactoptimizationlanguage} have been proposed to stabilize training or leverage multiple preferences. However, all remain within the direct preference optimization paradigm.

Traditional alignment methods primarily rely on reinforcement learning. RLHF \citep{ouyang2022traininglanguagemodelsfollow} and its variant RLAIF \citep{bai2022constitutionalaiharmlessnessai} optimize policies against learned reward models, requiring costly reward modeling and unstable online training, unsuitable under resource constraints.

SLiC \citep{zhao2023slichfsequencelikelihoodcalibration} reformulates alignment as supervised learning via critiques, while reward distillation approaches \citep{zhou2023limaalignment} simplify reward modeling yet still inherit RLHF inefficiencies.

In general, methods that avoid explicit reinforcement learning, such as DPO, offer more practical paths under constrained computational budgets.

\section{Methods}

All models were fine-tuned on a 0.5B parameter LLM (Qwen2.5-0.5B-Instruct) for 3 epochs. We used a batch size of 8, gradient accumulation, a maximum sequence length of 1024, and fp16 mixed-precision training. Training was conducted on an EC2 g6e.xlarge instance (48 GiB VRAM) and inference was accelerated using vLLM \cite{kwon2023efficient}.

Having established the training setup and baseline evaluation procedures, we now introduce the novel alignment methods developed in this project. These methods are designed to address the specific challenges of aligning small language models and aim to improve their human preference scores through tailored loss functions.

\subsection{Adaptive Margin-Sigmoid Loss}
In this design of the \texttt{margin-$\sigma$} loss function, the main goal is to combine the stability of log-sigmoid loss with the margin idea from hinge loss. We want to make the model not only learn correctly but also keep a safe distance from the decision boundary. This helps improve both training stability and generalization.

At first, we tried to change based on DPO. And we tried to add a RELU on the original sigmoid to increase the non-linearity, which also corresponds with the intuition that activation function's role. But after the experiment, we found that in the small model case,a relatively simple gradient path may contribute more. That's why we tried RELU only later.

The design has three main steps:

First, we compute the basic log-sigmoid loss for positive and negative samples, we define their difference as
\begin{align*}
\Delta r &=  \ln\!\tfrac{\pi_\theta(y_w\mid x)}{\pi_{\mathrm{ref}}(y_w\mid x)}- \ln\!\tfrac{\pi_\theta(y_l\mid x)}{\pi_{\mathrm{ref}}(y_l\mid x)}.
\end{align*}
Then, we compute the positive and negative log-sigmoid losses:
\begin{align*}
\ell_+ &= -\log \sigma(\beta \cdot \Delta r), \\
\ell_- &= -\log \sigma(-\beta \cdot \Delta r),
\end{align*}
where \(\sigma(\cdot)\) is the sigmoid function, and \(\beta\) is a scaling factor that controls how sharply the model separates preferred and rejected outputs. Using the log-sigmoid function ensures numerical stability even for very large or very small logits~\cite{NIPS2013_9aa42b31}.

Second, we introduce a margin by applying a ReLU cut-off. We subtract a small margin value \(\delta\) from both \(\ell_+\) and \(\ell_-\), and then apply the operator \([z]_+ = \max(0, z)\) to keep only the positive parts. Then, we combine the positive and negative losses using a small label smoothing coefficient \(\epsilon\), resulting in the final loss:
\begin{align*}
\mathcal{L}_{\text{margin}-\sigma} = (1-\epsilon)[\ell_+ - \delta]_+ + \epsilon[\ell_- - \delta]_+.
\end{align*}
\subsection{APO-hinge-zero}

Let's look back the equation of the \emph{Anchored Preference Optimization} objective in its zero–reference form \citep{doosterlinck2024anchoredpreferenceoptimizationcontrastive},
\begin{equation}
\mathcal{L}_{\text{APO-zero}}(x,y_w,y_l;\theta)= \hspace{4em}
\end{equation}
\[        
    \hspace{4em}-\sigma\!\bigl(r_\theta(x,y_w)\bigr)
    +\sigma\!\bigl(r_\theta(x,y_l)\bigr),
\]
\begin{equation}
\qquad
r_\theta(x,y)=\beta\log\frac{\pi_\theta(y\mid x)}{\pi_{\text{ref}}(y\mid x)},
\end{equation}
which adjusts the absolute likelihood of the winning and losing answers independently, making it naturally suited for \textbf{weak} LLMs whose initial probabilities are often indistinguishable from the reference model.
Empirically, however, we observe the standard logistic surrogate $\sigma(z)=\tfrac1{1+e^{-z}}$ continues to back-propagate non-negligible gradients even when a sample is already well classified, wasting computation and slowing convergence on small models.

To address this, we replace the logistic terms with a hard–margin hinge, with $[z]_{+} = \max(0,z)$:
\begin{equation}\label{eq:APO-hinge-zero}
\mathcal{L}_{\text{APO-hinge-zero}}(x,y_w,y_l;\theta)=\hspace{4em}
\end{equation}
\[
[m-\beta\,r_\theta(x,y_w)]_+
+
[m+\beta\,r_\theta(x,y_l)]_+,
\]
where $m$ is a fixed margin (we use $m\!=\!1$) and $\beta = 0.1$ retains its role as an inverse temperature.
The hinge keeps the same directional incentives—\emph{push} the winning answer up, \emph{pull} the losing answer down—but sets the gradient to zero once the condition
$|r_\theta(x,y)| \ge m/\beta$ is met.
This sparsifies updates, focuses compute on truly ambiguous pairs, and—crucially for 0.5 B-parameter models—prevents early gradient saturation. In preliminary experiments (Sec.~\ref{sec:results}), this method delivers the best AlpacaEval win rate (both Standard and Length-Controlled).

\subsubsection{Softplus-Hinge Loss}

During initial training experiments, we observed that due to the limited representational capacity of our smaller (0.5B parameters) model, the log-ratios rarely surpassed the predefined margin, even at later training stages. Specifically, using the original hinge-based objective (i.e., ReLU hinge),
most samples remained within the linear (positive) region of the ReLU function throughout the training process. With a small inverse temperature ($\beta=0.1$), the hinge objective behaved similarly to the original sigmoid-based loss, lacking the intended margin-driven gradient sparsification. Conversely, choosing a higher value for $\beta$ (e.g., $\beta\geq 1.0$) resulted in early and overly aggressive sparsification, significantly reducing gradient availability after just a few training iterations, causing premature convergence or stagnation of the model.

To address this challenge, we propose a smoothed hinge objective using the Softplus function, compactly defined as $s(x)=\log(1+e^x)$, which smoothly approximates the ReLU function with continuous gradients across the entire input domain. The modified objective thus becomes:
\begin{equation}
\mathcal{L}^{s}(x,y_w,y_l;\theta) 
=
\end{equation}
\[
 s(m - \beta\,r_\theta(x,y_w)) 
+ s(m + \beta\,r_\theta(x,y_l)).
\]

Unlike ReLU, which abruptly nullifies gradients once the input exceeds the margin, Softplus provides an exponentially decaying gradient even when significantly above or below the margin. This feature ensures stable, albeit smaller, gradient signals throughout training, preventing premature gradient starvation. 
    



\section{Results}
\label{sec:results}

We fine-tuned all models on the 0.5B parameter Qwen2.5-0.5B-Instruct model for three epochs, using a batch size of 8, gradient accumulation, a maximum sequence length of 1024, and fp16 mixed-precision training. Training was conducted on an EC2 g6e.xlarge instance, and inference was accelerated using vLLM~\cite{kwon2023efficient}. The total fine-tuning process took approximately 8 hours per model.

We evaluate a range of alignment methods on this setup, including several existing techniques and the novel approaches proposed in this project. Our experiments aim to benchmark the performance of different methods under a consistent training environment, with a particular focus on the challenges associated with small LLM alignment.

\subsection{AlpacaEval Results}
AlpacaEval is an automatic evaluation framework for instruction‐following language models. It leverages a strong LLM to provide pairwise preference judgments between a candidate model’s response and a baseline, thereby simulating human annotators at low cost and high speed. In experiments on 805 examples, AlpacaEval \cite{alpaca_eval} achieves high agreement with human labels (Spearman’s \(\rho \approx 0.94\)) . We \textbf{tested} our models on the full \textbf{AlpacaEval 1.0} set (805 prompts)

\begin{figure}[htbp]       
  \centering                
  \includegraphics[width=0.5\textwidth]{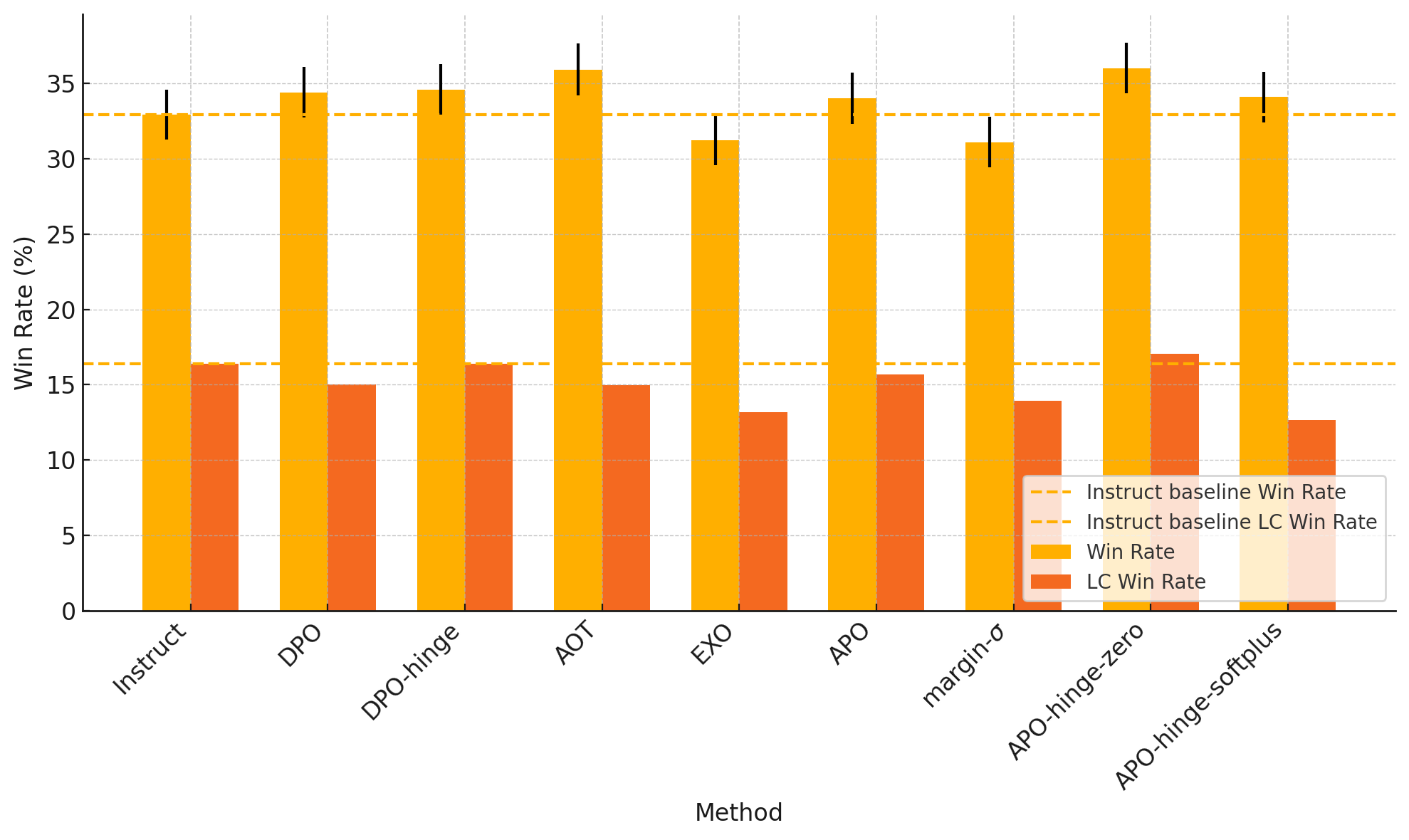}
  \caption{Alpaca Eval Win Rate of Alignment Methods (Tested on the whole Alpaca Eval Dataset with gpt4-turbo as a judge, compared with test-davinci-003)}
  \label{fig:alpacaeval}      
\end{figure}

Length‐Controlled AlpacaEval extends this approach by fitting a logistic regression that conditions on (1) the instruction, (2) the model identity, and (3) the difference in output length between the candidate and baseline. By evaluating the model at a zero length‐difference counterfactual, it reports an “LC Win Rate” that controls for length bias. This adjustment increases correlation with human judgments from 0.93 to 0.98 and reduces the ability to game the metric via output length by two‐thirds\cite{dubois2024length}.  We tested our models on the whole AlpacaEval dataset (805 prompts). As is shown in Figure~\ref{fig:alpacaeval}, the standard deviation of all of the models are around 1.7. 

Table~\ref{tab:alpacaeval} summarizes the numeric results extracted from Figure~\ref{fig:alpacaeval}.The most salient findings are:
\begin{itemize}
  \item \textbf{Highest Overall Win Rate}: APO-hinge-zero and AOT (both $\approx36\%$).
  \item \textbf{Highest LC Win Rate}: APO-hinge-zero ($\approx17\%$), indicating robustness under length control.
  \item \textbf{Underperformers}: EXO and margin-$\sigma$, both falling below the baseline in both metrics.
\end{itemize}

\begin{table}[ht]
  \centering
  \caption{AlpacaEval: Win Rate and LC Win Rate for each method (\%).}
  \label{tab:alpacaeval}
  \begin{tabular}{@{}lcc@{}}
    \toprule
    Method                & Win Rate & LC Win Rate \\
    \midrule
    APO-hinge-zero        & 36.02      & 17.07         \\
    AOT                   & 35.92      & 14.96         \\
    DPO-hinge             & 34.60      & 16.37         \\
    DPO                   & 34.41      & 15.02         \\
    APO-hinge-Softplus    & 34.10      & 12.67         \\
    APO                   & 34.02      & 15.68         \\  
    Instruct (baseline)   & 32.92      & 16.39         \\
    EXO                   & 31.22      & 13.15         \\
    margin-$\sigma$          & 31.11      & 13.91         \\
    \bottomrule
  \end{tabular}
\end{table}
Overall, the majority of alignment methods improve the baseline model’s win rate on AlpacaEval, yet many concurrently suffer reductions in LC win rate. This divergence underscores the length exploitation issue, whereby GPT’s preference for longer responses inflates perceived quality. Notably, APO-hinge-zero distinguishes itself by not only achieving the highest win rate but also enhancing the LC win rate, suggesting effective mitigation of the baseline model’s length exploitation. It should be noted that all models were evaluated under the same system prompt, thereby eliminating prompt variability as a confounding factor.

\subsection{MT-Bench Results}

MT‐Bench is a challenging multi‐turn evaluation benchmark consisting of 80 high‐quality question pairs designed to assess a chat assistant’s instruction‐following, coherence, and conversational ability over two turns. MT-Bench recently is often evaluated by strong LLMs like GPT-4 act as judges to compare model responses against reference answers, achieving over 80\% agreement with human preferences\cite{zheng2023judgingllmasajudgemtbenchchatbot}. MT-Bench thus provides a robust mechanism to gauge alignment effectiveness, particularly for small-scale models.

\begin{figure}
    \centering
    \includegraphics[width=1\linewidth]{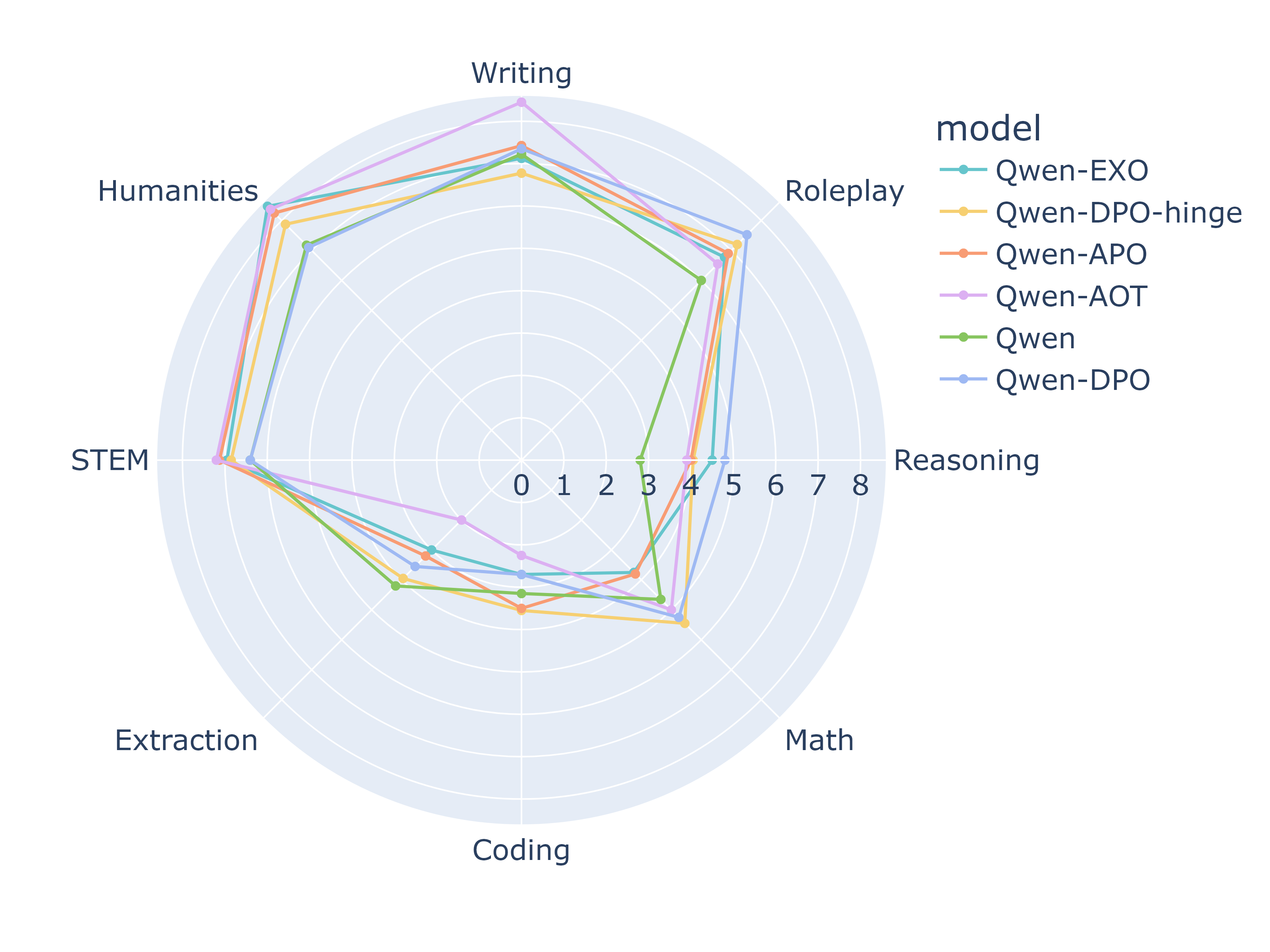}
    \caption{MT-bench Radar of Existing Methods}
    \label{fig:mtbench1}
\end{figure}
Figure~\ref{fig:mtbench1} summarizes the MT-Bench evaluation results across all tested alignment methods. In general, alignment enhances the model’s human-likeness and contextual responsiveness, translating effectively into performance gains on MT-Bench, notably within Writing, Roleplay, Reasoning, and Humanities categories. Conversely, limited or no improvements are observed in Extraction and Coding, alongside modest gains in Math and STEM, reflecting an inherent trade-off between precise task-specific correctness and conversational expressiveness.
We further analyze each alignment method individually to understand their respective strengths and weaknesses:
\begin{itemize}
\item \textbf{AOT:} Achieves the highest combined scores on Alpaca Eval and MT-Bench, marking it as the leading small-LLM alignment approach. Its exceptional performance in Writing significantly surpasses other methods, complemented by strong scores in STEM and Humanities. It sacrifices performance in these Extraction and Coding to achieve enhanced generative and expressive capabilities.

\item \textbf{DPO-hinge:} Provides the strongest performance in Math, alongside balanced improvements in Roleplay and STEM categories. Notably, it is among the few methods maintaining baseline-level performance in Extraction and Coding tasks. Despite minor regressions in Writing, DPO-hinge ranks second overall in Alpaca Eval, showcasing its well-rounded effectiveness.

\item \textbf{DPO:} Excels primarily in Roleplay and Reasoning, demonstrating consistently robust Math scores. Its balanced skillset underscores the method's versatility, particularly in conversational contexts.

\item \textbf{APO:} Demonstrates a balanced profile, exhibiting strong performance across Humanities, STEM, Coding, Roleplay, and Writing, thus reflecting an effective general-purpose alignment capability. 

\item \textbf{EXO:} Although achieving competitive scores within MT-Bench, the relatively low performance on Alpaca Eval restricts its practical utility and excludes it from further consideration within this context.
\end{itemize}

The results discussed above provide essential context and benchmarks against which the novel alignment methods proposed in this work will be evaluated and discussed.

\begin{figure}
    \centering
    \includegraphics[width=1\linewidth]{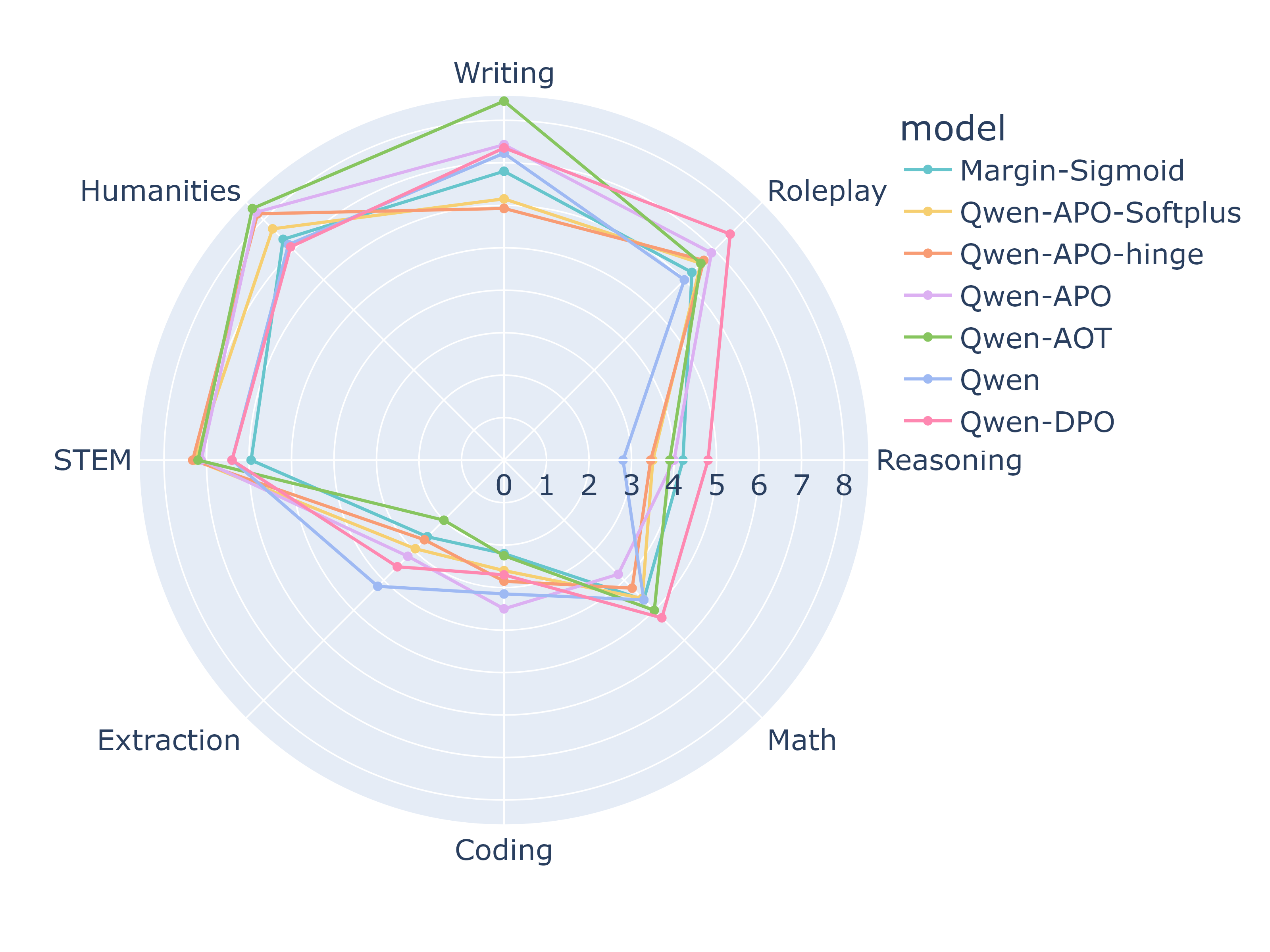}
    \caption{MT-bench Radar comparing Competitive Existing Methods and Our methods}
    \label{fig:mtbench2}
\end{figure}

This project proposes two novel alignment objectives: Margin-Sigmoid, APO-hinge-zero, and a Softplus variant APO-hinge-Softplus.

\begin{itemize}
    \item \textbf{Margin-Sigmoid}: This method performs relatively poorly on AlpacaEval but achieves a more balanced performance across different categories in MT-Bench.
    
    \item \textbf{APO-hinge-zero}: This method achieves the best performance on AlpacaEval. However, its performance on MT-Bench is less impressive. Specifically, it excels in STEM-related tasks and performs well in Humanities, but exhibits only average performance in other domains. 
    
    \item \textbf{APO-hinge-Softplus}: Overall, this method performs similarly to APO-hinge-zero. Notably, it shows slightly better performance than APO-hinge-zero in non-STEM, non-Humanities categories on MT-Bench.
\end{itemize}

Based on these observations, we now discuss several key points.

\section{Discussion}

\subsection{Hinge Variants Outperform Sigmoid Counterparts on AlpacaEval}
\label{sec:hinge_background}

\paragraph{Empirical background.}
Across all experiments on AlpacaEval, objectives that replace the
sigmoid margin with a hinge margin consistently yield higher scores.
\begin{itemize}
    \item \textbf{DPO$\rightarrow$DPO--Hinge.}
    Win-rate $+0.2$\,pts; Length-controlled (LC) win-rate $+1.35$\,pts.
    \item \textbf{APO$\rightarrow$APO--Hinge--zero.}  
    Win-rate $+2.0$\,pts; LC win-rate $+1.39$\,pts.
    \item \textbf{Soft variants.}  
    APO-hinge-Softplus ---which smooth the cutoff with
    \texttt{Softplus} ---deliver only
    marginal gains, suggesting that the \emph{hard} threshold is essential. Also, the $\beta$ for APO-hinge-Softplus was set too high.
\end{itemize}
These observations motivated us to transplant the hinge formulation from DPO
into structurally similar objectives.  The resulting APO-hinge-zero achieves the
highest overall score in our study.

\paragraph{An intriguing side-effect.}
Besides boosting raw performance, the hinge objectives also improve
\textit{LC win-rate}, a metric that removes the advantage conferred by longer
answers.  In other words, hinge training appears to attenuate the notorious
\emph{length exploitation} bias.  Why does a seemingly simple change in the
loss function regularise the model in this manner?

\subsection{Hard-Example Mining as an Implicit Regulariser}
\label{sec:hinge_mechanism}

We posit that the answer lies in the hinge loss’s built-in \emph{hard-example
mining}.  For a preference pair $(x,y_w,y_\ell)$ with log-ratio reward
$r_\theta=\log\pi_\theta-\log\pi_{\mathrm{ref}}$, the hinge objective is
\[
\mathcal{L}_{\text{hinge}}
=\bigl[\,m-\beta\bigl(r_\theta(x,y_w)-r_\theta(x,y_\ell)\bigr)\bigr]_+ .
\]
For APO-hinge-zero, it is Equation~\ref{eq:APO-hinge-zero}. Pairs where the chosen answer already enjoys a large margin
($r_w-r_\ell>m/\beta$) incur zero loss and emit no gradient.  Because
many 'easy' pairs in AlpacaEval favor the longer answer,
these length-driven examples saturate early in training.  All subsequent
updates therefore concentrate on \emph{hard} pairs where length alone
is insufficient, forcing the model to improve coherence and factuality
instead of merely inflating output length.  The observed rise in LC
win-rate is a direct consequence of this gradient reallocation. 

\paragraph{Future directions.}
We plan to (i) quantify the active-pair fraction across length buckets to
verify that long-answer pairs saturate first; (ii) sweep $\beta$ (especially for APO-hinge-Softplus) and the
margin $m$ to map the trade-off between bias suppression and convergence
speed; and (iii) compare hinge-based hard-example mining with explicit
length-penalty regularisers and curriculum scheduling.  Establishing causal
links between pair difficulty, gradient allocation, and human preference
ratings remains an open research avenue.

\subsection{Why AOT is Particularly Effective for Small LLMs}

Small LLMs are particularly sensitive to optimization noise due to their limited representational capacity. Traditional preference optimization methods such as DPO and DPO-Hinge focus on optimizing individual, per-sample preference margins. This local approach can introduce significant gradient noise, leading to instability or overfitting, especially when the distinctions between chosen and rejected responses are subtle or ambiguous.

In contrast, AOT optimizes alignment at a distributional scale (as detailed in Equation~\ref{eq:aot_objective}). Specifically, by minimizing the optimal transport cost defined in Equation~\eqref{eq:aot_ot_cost}, AOT explicitly enforces a global preference ordering, ensuring the chosen response distribution stochastically dominates the rejected distribution. This global constraint naturally smooths the optimization trajectory, significantly reducing the impact of local gradient fluctuations.

For small-scale models, such smoothing is critical because these models lack sufficient internal mechanisms (e.g., layers, attention heads) to effectively average out noisy gradients. Furthermore, the global FSD constraint (Eq.~\eqref{eq:aot_objective}) aggregates alignment signals across the entire data distribution, providing a stronger and more coherent training signal. This prevents the model from overfitting to noisy or individually ambiguous preference pairs, ultimately yielding better generalization performance on challenging benchmarks.

Therefore, the combination of a distributional alignment objective and the stable sorting-based optimization inherent in AOT significantly enhances the training effectiveness for small LLMs.

\subsection{Limitations of the \textit{Margin--Sigmoid} Loss}
\label{sec:margin_sigma_limits}

The \textit{Margin--$\sigma$} objective was motivated by an appealing intuition:  
combine the numerical stability of the log–sigmoid loss used in Direct Preference Optimisation (DPO) with the safety buffer provided by hinge‐style margins.  

In practice, however, the method delivers only modest gains over EXO and remains consistently inferior to DPO, AOT, and their hinge‐based derivatives.  
Below we dissect the failure modes in detail and outline several avenues for refinement.

\subsubsection{Gradient starvation in the early curriculum}

The defining operation of \textit{Margin--$\sigma$} is a \emph{hard loss truncation}:
\[
   \tilde{\ell}_\pm \;=\; \bigl[\,\ell_\pm-\delta\,\bigr]_+,
\quad
   \ell_\pm \;=\;-\log\sigma(\pm\beta\Delta r),
\]
which sets the gradient to zero whenever the log–sigmoid term already falls below a user-chosen threshold $\delta$.  Because small LLMs have limited capacity and cannot yet handle linguistically subtle examples, they depend critically on these \emph{easy pairs} to acquire basic syntax, factual priors, and stylistic conventions.  
By discarding them, the optimizer forces the network to tackle ambiguous or noisy pairs before it has learned the underlying ``common-sense’’ heuristics.

\subsubsection{Loss of fine-grained ranking resolution}

DPO and AOT continue to nudge the model even after it has learned the correct ordering, progressively calibrating the \emph{relative} strengths between competing responses.  
In contrast, once \(\ell_\pm<\delta\) the \textit{Margin--$\sigma$} objective treats every pair as equally solved.  
This coarse resolution prevents the model from refining subtle preference gaps, thereby capping attainable win rates several percentage points below those of DPO-hinge and AOT.

\subsubsection{Length Exploitation}

Another important issue, not unique to Margin\_$\sigma$ but shared across all RLHF-free alignment methods, is the phenomenon of length exploitation. Models fine-tuned with these methods tend to "cheat" by generating longer outputs, which are often superficially preferred by evaluators, even when true reasoning quality or factual correctness does not improve. Empirical results show that while raw win rates against text-davinci-003 outputs reach around 30\% to 35\%, the length-normalized win rates consistently fall to approximately 13\% to 16\%.

We hypothesize that this behavior is exacerbated in small models (e.g., models under 1B parameters). Due to limited capacity, small models struggle to learn complex patterns in the data and instead latch onto easily learnable but imperfect heuristics, such as "longer answers are better." As a result, most alignment methods, including Margin\_$\sigma$, inadvertently amplify this length bias, rather than mitigating it. Only a few approaches, such as DPO-hinge and APO-hinge-zero, show some ability to resist this exploitation effect by structurally modifying the loss to discourage over-reliance on answer length.

\subsubsection{Brief directions for improvement.}
Our ablations suggest three lightweight modifications that preserve the original intuition while addressing the issues above:

\begin{enumerate}
\item \textbf{Smooth truncation.}  
      Replace the hard ReLU \([\cdot]_+\) with a temperature-controlled Softplus,  
      \(\operatorname{Softplus}_\tau(z)=\tau\log\bigl(1+e^{z/\tau}\bigr)\),  
      or adopt a focal-style weight \(w(\ell)=(1-e^{-\ell})^\gamma\) that down-weights but does not extinguish easy pairs.

\item \textbf{Curriculum margin.}  
      Initialise training with \(\delta=0\) (i.e.\ vanilla DPO) and linearly raise it to the target value over the first 1–2 epochs, allowing the model to build a stable lexical basis before emphasising harder distinctions.

\item \textbf{Explicit anti-length penalty.}  
        Integrate anti-length regularization strategy such as \citep{park2024disentanglinglengthqualitydirect}
\end{enumerate}

\section{Conclusions}

\subsection{What we achieved}
We have introduced two novel alignment objectives for small‐scale LLMs—Adaptive Margin‐Sigmoid Loss and APO‐hinge-zero (with Softplus variant)—that blend margin‐based hard‐example mining with APO’s chosen‐focused optimization. By integrating the implicit regularisation effect of hinge-based hard-example mining with the chosen–focused optimisation signal of APO–zero, APO–hinge-zero simultaneously filters out trivial samples and concentrates learning on preferred outputs. This synergistic design underpins its strong and robust performance across both standard and length-controlled AlpacaEval win rates. On the other hand, the other two methods delivered MT-Bench results on par with competing models. These findings confirm both the soundness and feasibility of our proposed methods.

Empirically, APO‐hinge-zero achieves the best performance on AlpacaEval, posting a Win Rate of 36.02\% and an LC Win Rate of 17.07\%, while other variants match competing models on MT-Bench across STEM, Humanities, and dialogue tasks. These results confirm that targeted, lightweight modifications to preference‐optimization losses can markedly improve alignment quality and efficiency under resource constraints.

\subsection{Limitations}

For existing models, we mostly adopted industry-standard or originally recommended hyperparameter settings. For our newly proposed methods, we conducted only minimal-scale hyperparameter tuning based on intuition, due to time and resource constraints. Consequently, our reported results may carry some bias, and certain models might have been evaluated with suboptimal hyperparameter choices.

We also did not test across a broader range of benchmarks due to limited computational resources and budget. Additionally, we only evaluated a subset of alignment methods — specifically, those that are widely used in practice or are hypothesized to perform reasonably well on small models — rather than exhaustively benchmarking all existing approaches.

Compared to larger models (e.g., those achieving $\sim$90\% win rate on AlpacaEval 1.0), our 0.5B model achieves only $\sim$30\% win rate against \texttt{text-davinci-003}, with performance further decreasing after length normalization.

\section*{Code and Data Availability}
All code and evaluation scripts are available at \url{https://github.com/Oaynerad/Alignment-on-small-LLM}.

\bibliography{custom}

\appendix



\end{document}